\documentclass[10pt,twocolumn,letterpaper]{article}

\usepackage[pagenumbers]{cvpr} % To force page numbers, e.g. for an arXiv version

\usepackage{graphicx}
\usepackage{amsmath}
\usepackage{amssymb}
\usepackage{booktabs}
\usepackage{multirow}
\usepackage{colortbl}

\usepackage[pagebackref,breaklinks,colorlinks]{hyperref}

\usepackage[capitalize]{cleveref}
\crefname{section}{Sec.}{Secs.}
\Crefname{section}{Section}{Sections}
\Crefname{table}{Table}{Tables}
\crefname{table}{Tab.}{Tabs.}

\begin{document}

\title{An Empirical Study of End-to-End Temporal Action Detection}

\author{
Xiaolong Liu$^{1}$~~~~
Song Bai$^{2}$~~~~
Xiang Bai$^1$\thanks{Corresponding author}\\
$^1$Huazhong University of Science and Technology
$^2$ByteDance Inc. \\
{\tt\small \{liuxl, xbai\}@hust.edu.cn, songbai.site@gmail.com} \\
}

\maketitle

\begin{abstract}
Temporal action detection (TAD) is an important yet challenging task in video understanding. It aims to simultaneously predict the semantic label and the temporal interval of every action instance in an untrimmed video. Rather than end-to-end learning, most existing methods adopt a head-only learning paradigm, where the video encoder is pre-trained for action classification, and only the detection head upon the encoder is optimized for TAD. The effect of end-to-end learning is not systematically evaluated. Besides, there lacks an in-depth study on the efficiency-accuracy trade-off in end-to-end TAD. In this paper, we present an empirical study of end-to-end temporal action detection. We validate the advantage of end-to-end learning over head-only learning and observe up to 11\% performance improvement. Besides, we study the effects of multiple design choices that affect the TAD performance and speed, including detection head, video encoder, and resolution of input videos. Based on the findings, we build a mid-resolution baseline detector, which achieves the state-of-the-art performance of end-to-end methods while running more than 4$\times$ faster. We hope that this paper can serve as a guide for end-to-end learning and inspire future research in this field. Code and models are available at \url{https://github.com/xlliu7/E2E-TAD}.

\end{abstract}

%%%%%%%%% BODY TEXT
\section{Introduction}
\label{sec:intro}
With the development of information technology, the numbers of videos generated and accessed are rapidly increasing, underscoring the need for automatic video understanding, such as human action recognition and temporal action detection (TAD)\footnote{Also known as temporal action localization (TAL).}. Action recognition aims to predict the action label (~\eg, basketball dunk) of a short, trimmed video. Differently, TAD aims to determine the label, as well as the temporal interval of every action instance in a long untrimmed video. It is more challenging and also practical in real-world actions, such as security surveillance, sports analysis, and smart video editing.

\begin{figure}[t]
    \centering
    \includegraphics[width=0.9\linewidth]{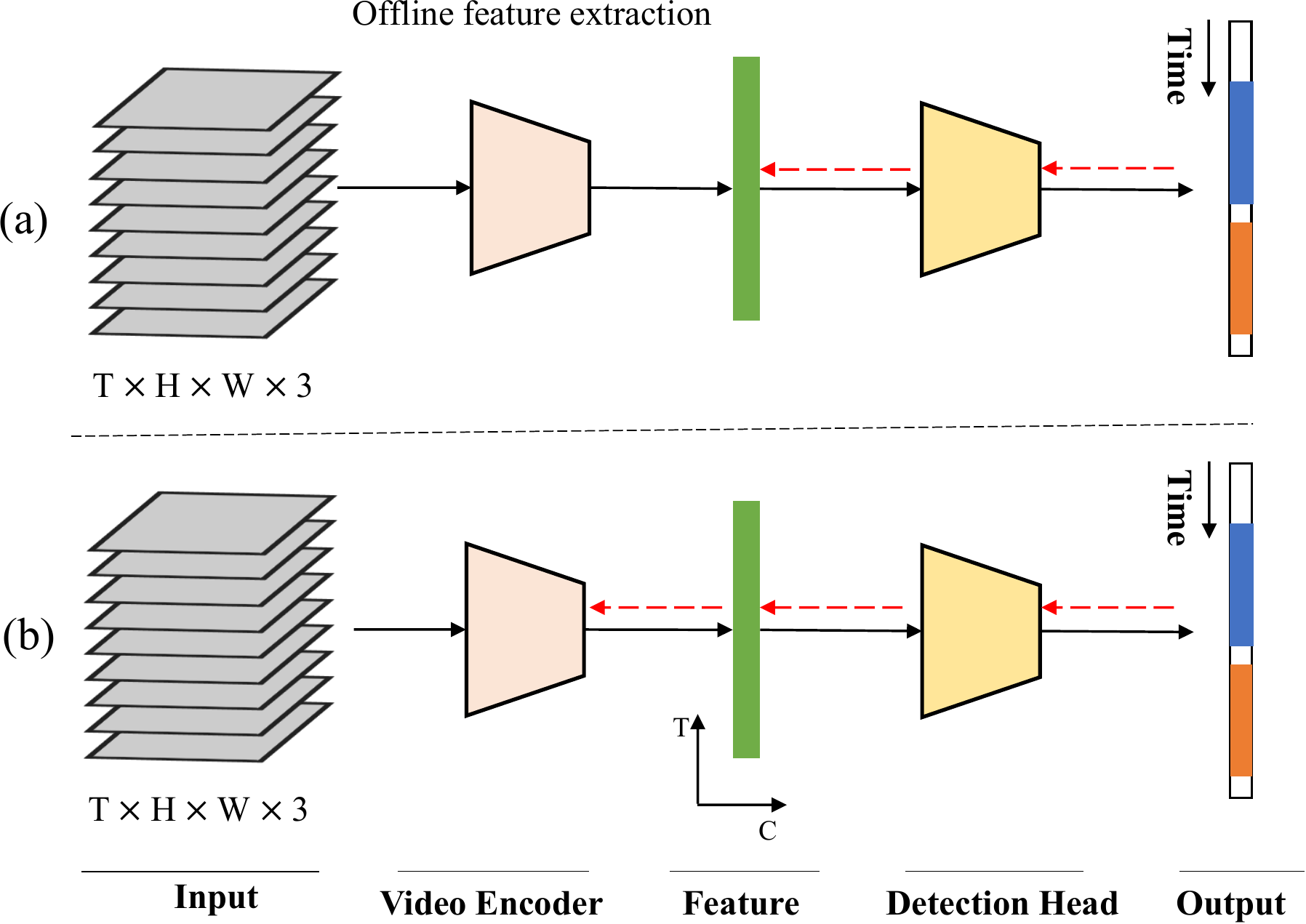}
    \caption{Head-only learning (a) \vs~ end-to-end learning (b) for temporal action detection. Solid arrows and dashed arrows represent forward pass and the gradient flow of back propagation.}
    \label{fig:intro}
\end{figure}

\begin{table}[]
    \centering
    \small

    \begin{tabular}{l|ccccc}
    \toprule
    Method &E2E&Flow&FLOPs&Latency&mAP\\
    \midrule
    \multicolumn{6}{c}{\textbf{THUMOS14}}\\
    \midrule
        MUSES~\cite{Liu_2021_CVPR}&&\checkmark&17.5T& 72s*+2.1s&53.4\\
        AFSD~\cite{lin2021learning}&\checkmark&\checkmark& 2780G&2472ms&52.0\\
        Ours&\checkmark&&475G&587ms&\textbf{54.2}\\
        \midrule
    \multicolumn{6}{c}{\textbf{ActivityNet}}\\
    \midrule
    AFSD~\cite{lin2021learning}&\checkmark&\checkmark& 499G&291ms&34.39 \\
    Ours&\checkmark&&62G&63ms&\textbf{35.10}\\
    \bottomrule
    \end{tabular}
        \caption{Comparison between the baseline detector built in this work (ours) with state-of-the-art methods. The latency and FLOPs are measured at the video level. The time of optical flow extraction is not included in latency. *The time cost of I3D~\cite{carreira2017quo} feature extraction. E2E: end-to-end.
    % For reference, the average length per video is 4.4 minutes and 2 minutes on THUMOS14 and ActivityNet.
    }
    \label{tab:runtime}
\end{table}

Owing to the strong discriminative power of neural networks, deep learning methods have dominated the field of temporal action detection~\cite{lin2019bmn,xu2020g,zeng2019graph,yang2021background}. 
As depicted in Fig.~\ref{fig:intro}, a temporal action detector typically consists of a \textbf{video encoder} and a \textbf{detection head}, similar to the backbone-head structure in object detection~\cite{ren2015faster,tian2019fcos,Girshick_2015_ICCV}. 
Different from modern object detectors that are trained end-to-end\footnote{ ``End-to-end'' has diverse meanings in literature. Here we mean joint learning of the video encoder and the detection head in a detector.}, 
most TAD methods adopt a \textbf{head-only learning} paradigm.
They first pre-train the video encoder on a large action recognition dataset (\eg, Kinetics~\cite{carreira2017quo}) then freeze it for offline feature extraction. 
After that, only the detection head upon the features is trained for the TAD task on the target datasets. This leaves the video features sub-optimal and restricts the performance.

Although a few works~\cite{xu2017r,lin2021learning,long2019gaussian} have adopted end-to-end learning,
there lacks an in-depth analysis of it. The actual benefit of end-to-end learning is still unclear.
Besides, the effects of many factors in end-to-end TAD, such as the video encoder, the detection head, the image and temporal resolution of input videos, are not systematically studied. In a way, lack of such a study blocks the research of end-to-end TAD.
Moreover, existing works more or less neglect the efficiency, which is an important factor in real-world applications. For example, in large-scale systems, such as online video platforms, running time determines computational expenses. 
Unfortunately, most methods do not discuss the computation cost. 
A few works discuss the running time of certain parts of the full model, \eg, the detection head ~\cite{lin2019bmn,qing2021temporal,zhao2021video,liu2021end} or report inference speed (FPS)~\cite{lin2021learning,xu2017r}. But they do not explore the efficiency-accuracy trade-off. 
This paper aims to address the above issues.

We conduct an empirical study of end-to-end temporal action detection. 
Four video encoders and three detection heads with different high-level designs are evaluated on two standard TAD datasets,~\ie, THUMOS14 and ActivityNet. \textbf{Firstly}, we uncover the benefit of end-to-end learning.
It is shown that end-to-end trained video encoders with a medium image resolution ($96^2$) can match or surpass pre-trained ones with standard image resolution ($224^2$) in terms of TAD performance. \textbf{Secondly}, we evaluate the effect of a
series of design choices on performance and efficiency, including detection head, video encoder, image resolution and temporal resolution.
It may serve as a guide for seeking the efficiency-accuracy trade-off.
\textbf{Lastly}, we build a baseline detector based on our study. It achieves state-of-the-art performance of end-to-end TAD while running more than 4$\times$ faster (see Tab.~\ref{tab:runtime}). 
Specifically, it can process a 4-minute video in only 0.6 seconds. 
We hope that our work can facilitate future research in temporal action detection.

\section{Related Works}
\noindent\textbf{Temporal Action Detection Methods.} 
Current temporal action detection methods can be roughly categorized into
three groups.
\textbf{Anchor-based methods}~\cite{zeng2019graph,zhao2017temporal,chao2018rethinking,lin2021learning,qing2021temporal,zhu2021enriching,sridhar2021class,li2021three} first generate a dense set of anchors,~\ie, temporal segments that may contain an action, then leverage a classifier to classify them into background or one action class. 
In these methods, anchors are generated by uniform sampling~\cite{buch2017sst, chao2018rethinking, escorcia2016daps, gao2017turn, shou2016temporal,xu2017r}, grouping potential action boundaries~\cite{lin2018bsn, shou2017cdc,zhao2020bottom, zhao2017temporal}, or a combination of the them~\cite{liu2019multi,gao2018ctap}. 
\textbf{Anchor-free methods}~\cite{lin2017single,buch2017end,lin2021learning,shou2017cdc,yuan2017temporal} directly predict the action class for each frame in the video. Then they group frames with the same class into temporal segments.
Some methods~\cite{lin2021learning,yang2020revisiting} additionally regress the distance to action boundaries.  
\textbf{Query-based methods}~\cite{tan2021relaxed,liu2021end} draw inspiration from the DETR object detection framework~\cite{carion2020end}.
They take as input a small set of learnable embeddings called action queries and video features, and map each query to an action prediction. This is achieved via Transformer attention~\cite{vaswani2017attention} that models the relations between query embeddings and video features. 
Owing to a one-to-one matching mechanism between ground truth actions and queries, they generate spare and unique action predictions.
Different from previous methods that mostly focus on the design of network architecture or framework, we focus on the learning paradigm and efficiency-accuracy trade-off.

\vspace{0.8ex}\noindent\textbf{Video Encoders.}
The video encoders in TAD are adapted from action recognition networks by dropping the classification heads.
In previous methods, two-stream networks (\eg, TSN~\cite{wang2016temporal}) and 3D Convolutional Neural Networks (\eg, C3D~\cite{tran2015learning}, I3D~\cite{carreira2017quo}) are commonly used video encoders.
Two-stream networks, firstly proposed in~\cite{simonyan2014two}, consist of two 2D Convolutional Neural Network (CNN) streams that operate on RGB frames and optical flow frames separately and their outputs are fused. In two-stream methods, optical flow is crucial for high performance as they explicitly capture motion cues. However, the calculation of optical flow is very expensive.
Differently, 3D networks can capture motion information from a sequence of frames, at the cost of more parameters and computation than 2D networks. 
I3D~\cite{carreira2017quo}, a representative of this kind, is widely used in previous TAD methods.
To mitigate the above issues of 3D networks, recent methods~\cite{xie2018rethinking,tran2018closer,qiu2017learning,lin2019tsm,feichtenhofer2019slowfast} use different ways to approximate 3D convolution. For example, decomposing 3D convolution into 1D and 2D convolution, or combining a temporal shift operation ~\cite{lin2019tsm} with 2D convolution. 
In this paper, we evaluate various video encoders to examine their performance and efficiency in temporal action detection. Their effects have not been systematically studied before.

\vspace{0.8ex}\noindent\textbf{Learning Paradigms of TAD.} Most TAD methods first extract features with video encoders pre-trained on action recognition (classification) datasets (\eg~Kinetics-400~\cite{carreira2017quo}, similar to the role of ImageNet in image recognition). Then they train and evaluate the detection head with the extracted features.  In this way, the experimental period can be greatly shortened. Therefore, it is adopted by most existing works. However, there are two issues in this learning paradigm, task inconsistency and data inconsistency between the pre-training stage and the downstream TAD stage. 
To deal with the task inconsistency issue,~\cite{xu2021boundary} designs a pre-training task that classifies synthesized video clips with different kinds of boundaries.
To cope with the data inconsistency issue, some works~\cite{long2019gaussian,piergiovanni2019temporal,alwassel_2021_tsp} pre-train the video encoder for action recognition on the target TAD dataset. 
This paper explores an alternative way of end-to-end training on the TAD datasets. The goal of this paper is not to compare end-to-end training with other pre-training techniques. Instead,
we aim to explore the effects of a series of factors on speed and accuracy and seek a trade-off between them.

\section{Experimental Setup}
In this section, we review the video encoders and temporal action detection heads that we study in this paper. The datasets for performance evaluation and the implementation details are also described here.

\subsection{Video Encoders}
\begin{figure}
    \centering
\includegraphics[width=\linewidth]{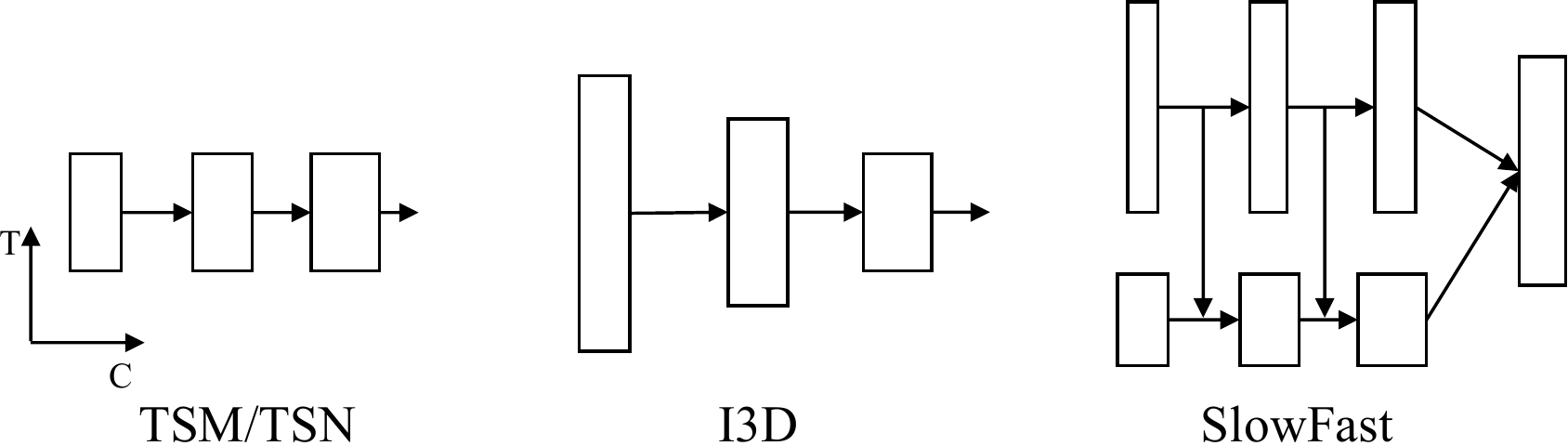}
    \caption{A high-level diagram of the video encoders studied in this work. For simplicity, we do not show the spatial dimension.}
    \label{fig:encoder_fig}
\end{figure}

We mainly study four kinds of video encoders, TSN~\cite{wang2016temporal}, TSM~\cite{lin2019tsm}, I3D~\cite{carreira2017quo} and SlowFast~\cite{feichtenhofer2019slowfast}. Fig.~\ref{fig:encoder_fig} illustrates the network structures of these video encoders. 

\vspace{1ex}\noindent\textbf{TSN} is a pure 2D CNN encoder. It processes each frame independently.

\vspace{1ex}\noindent\textbf{TSM} combines a temporal shift operation with 2D convolution as a basic building block of video encoders. The shift operation moves a small fraction of channels of the input feature map forward and another fraction backward in the temporal axis. It is equivalent to temporal 1D convolution with constant parameters but introduces no computation cost. Spatiotemporal features from multiple frames are then captured with 2D convolution on the shifted features.

\vspace{1ex}\noindent\textbf{I3D} follows the design of the Inception network~\cite{DBLP:conf/icml/IoffeS15} for image recognition but inflates all convolutional and pooling layers into 3D counterparts. As temporal pooling is involved, it outputs feature maps with different resolution in different stages of the network. 

\vspace{1ex}\noindent\textbf{SlowFast} (SF) consists of a slow pathway and fast pathway that operate on sparsely and densely sampled video frames respectively. The fast pathway has fewer channels than the slow pathway. Therefore it can efficiently capture motion information, which is fused to the slow pathway stage by stage. It follows recent works~\cite{qiu2017learning,tran2018closer} to apply 1D and 2D convolution iteratively.

\subsection{Temporal Action Detection Heads}
We study three kinds of temporal action detection heads (methods), anchor-based, anchor-free, and query-based.
% . An exhaustive evaluation of all existing TAD heads are unrealistic. Therefore
G-TAD~\cite{xu2020g}, AFSD~\cite{lin2021learning}, and TadTR~\cite{liu2021end} are selected as the representative of each kind for their state-of-the-art performances. Here we briefly describe their frameworks.

\vspace{1ex}\noindent\textbf{G-TAD} views a video as a graph and all snippets in the video as its nodes. With such a formulation, the context information in the video can be captured by graph convolution on these nodes. These nodes are sampled as potential action boundaries and paired nodes become anchors. Similar to RoIAlign~\cite{he2017mask}, an SGAlign operation is designed to extract aligned features within the temporal region of each anchor. These anchors are then classified by several fully connected layers upon the aligned features.

\vspace{1ex}\noindent\textbf{AFSD} is an anchor-free detector. Inspired by the anchor-free methods~\cite{tian2019fcos,qiu2020borderdet} in object detection, it detects actions by predicting the action class and the distances to action boundaries for each frame. Using this formulation, it first generates coarse action predictions with pyramid features from the video encoder. To enhance the detection performance, a saliency-based refinement module is designed. It extracts the salient features around the boundaries of each predicted action via a boundary pooling operation. These features are utilized to generate refined predictions.

\vspace{1ex}\noindent\textbf{TadTR} views TAD as a direct set prediction problem. Based on Transformer~\cite{vaswani2017attention}, it maps a small set of learned action query embeddings to corresponding action predictions with a Transformer encoder-decoder architecture. The Transformer encoder takes as input the features from the video encoder. It models the long-range dependency in the temporal dimension with a sparse attention mechanism and captures the global context. The decoder looks up global context related to each query via cross-attention and predicts the boundaries and the action class thereon. 
In pursuit of more accurate boundaries and confidence scores, it utilizes a segment refinement mechanism that iteratively refines the boundaries in each decoder layer and an actions regression head that re-computes a confidence score according to the final predicted boundaries.

% By default, we use the TadTR detection head for its simplicity and efficiency.

\subsection{End-to-end Learning}
We drop the classifier in the original network of each video encoder and modify the last global pooling layer to only perform spatial pooling.
Then the detection head is attached to the last layer of the encoder, resulting in a unified network. The network directly takes video frames as input and is trained with the loss functions defined by each detector. During training, gradients flow backward to both the head and the video encoder. In this way, they can be optimized simultaneously towards stronger temporal action detection performance.

\subsection{Datasets}
We conduct evaluations on two datasets, THUMOS14~\cite{jiang2014thumos} and ActivityNet\cite{caba2015activitynet} (v1.3).
\textbf{THUMOS14} collects sports videos from 20 classes. It contains 200 and 212 untrimmed videos for training and testing. The actions are densely distributed and very short.
The average length of videos and actions is 4.4 minutes and 5 seconds respectively. 
\textbf{ActivityNet} consists of 19994 videos in 200 action classes of daily activities. It contains 10024, 4926, and 5044 videos in the training, validation, and testing sets.
Following previous work, we use the validation set for evaluation, as the annotations on the testing set are reserved by the organizers. The average length of videos and actions is 2 minutes and 48 seconds respectively. 

\vspace{1ex}\noindent\textbf{Evaluation Metrics.} For both datasets, we use mean Average Precision (mAP) at different temporal IoU thresholds as the evaluation metric. On THUMOS14, the IoU thresholds are $\{0.3, 0.4, 0.5, 0.6, 0.7\}$. On ActivityNet, we choose 10 values ranging from 0.5 to 0.95 with a step of 0.05. We also report the average of the mAP at all thresholds, which is the primary metric for performance comparison. 

\subsection{Implementation Details}
\noindent\textbf{Video Encoders.} The SlowFast encoder has several variants. We choose the ``SlowFast 4x16, R50" variant for its efficiency. Given an input clip of N frames, the fast and the slow pathway sample N and N/8 frames respectively. We resize the output features of the two pathways to the same length and concatenate them into one. The length is set to N/4. In other words, the temporal output stride is 4.
I3D extract features of multiple temporal resolutions. A \textbf{feature fusion} strategy is applied to better utilize these features. We temporally up-sample the features from the fifth stage by $2\times$ and fuse it with the features from the fourth stage. In this way, the temporal output stride is also 4. As a reference, the temporal output stride of TSM and TSN is 1.

\vspace{1ex}\noindent\textbf{Clip Sampling.} 
We use video clips for training and evaluation. On THUMOS14, we uniformly sample clips of 25.6 seconds, which is longer than 99.6\% of all action instances. The sampling stride between adjacent clips is set to 25\% and 75\% of the clip length during training and evaluation, respectively. Unless specially noted, TSM and TSN sample video frames at 3.75 FPS on THUMOS14. SlowFast and I3D sample frames at 10 FPS.  On ActivityNet, as the ratio of action length to video length is much larger, we follow~\cite{lin2021learning} to treat each full video as a single clip and sample a fixed number of frames as the input to video encoders. According to~\cite{lin2021learning}, this strategy is better than sampling with a fixed frame rate.
This number is set to 384 for SlowFast and I3D and 96 for TSM and TSN. In this way, the output features of these encoders have the same length of 96 (an average of 0.8 FPS). 
By default, we set the image size of the input video to 96$\times$96, which has 5.4$\times$ fewer pixels than the commonly used 224$\times$224 resolution. 

\vspace{1ex}\noindent\textbf{Training.} The models are trained with Adam~\cite{kingma2015adam} optimizer, setting weight decay to $10^{-4}$. The base learning rate is set to $10^{-4}$ and $5\times 10^{-5}$ on THUMOS14 and ActivityNet empirically.  The learning rate of the video encoder is multiplied by a factor of 0.1, which helps to stabilize training. We divide the learning rate by 10 after $\tau_1$ epochs and the total number of epochs is $\tau_2$. We set $\tau_1=10$ and $\tau_2=12$ on THUMOS14. On ActivityNet, they are set to $8$ and $10$, respectively. We set the batch size to 4 for all models and freeze the batch normalization layers in the video encoders. 
With this configuration, most models can be trained using \textbf{a single GPU with 12 GB of memory}. We analyze the effect of batch size in the supplementary and observe that varying batch size from 4 to 16 gives similar performance. We use cropping, horizontal flipping, rotation and photometric distortion for image augmentation. 
The angle range of random rotation is (-45, 45) degree. The settings of photometric distortion follow~\cite{liu2016ssd}. The probability of the latter three transformations is 0.5.

\vspace{1ex}\noindent\textbf{Inference.} We follow the details of each detection head in their original implementation. 
On ActivityNet, we follow previous works~\cite{lin2018bsn,lin2019bmn,zhao2021video,lin2021learning,xu2020g,long2019gaussian,xu2021low} to perform class-agnostic localization and use the video-level classification labels from~\cite{zhao2017cuhk}.
Latency is measured on a single TITAN Xp GPU, with the batch size set to 1. We take the average time of 100 runs after 10 warm-up runs. Unless specially noted, the computation costs on THUMOS14 are measured for video clips of 25.6 seconds.

\section{Results and Analyses}

\subsection{The Effect of End-to-end Learning}
\noindent\textbf{Head-only~\vs~E2E.} In Tab.~\ref{tab:finetuning}, we compare the performance of traditional head-only learning and end-to-end learning using the TadTR detector.
When studying the performance gain of end-to-end learning, we keep the same mid-resolution (96 $\times$ 96) setting.
We also list the performance of head-only learning with 224$\times$ 224 resolution. We see that:

\noindent(I) End-to-end learning consistently improves performance on multiple datasets and backbones. On THUMOS14, end-to-end learning improves the average mAP by 9.41\% and 11.21\% with TSM ResNet-18 and TSM ResNet-50 encoders respectively. On ActivityNet, it achieves an improvement of 1.30\% and 1.38\% average mAP with the two encoders respectively. We show that this also generalizes to other video encoders (I3D and SlowFast) and detection heads (AFSD and G-TAD) \textbf{in the supplementary}.

\noindent(II) The performance of mid-resolution ($96^2$) end-to-end models can match or surpass that of standard-resolution ($224^2$) models trained in the head-only paradigm. On THUMOS14, the former outperforms the latter by 7.52\% (45.25\% \vs~37.77\%) on the TSM ResNet-50 encoder. A similar observation is drawn on TSM ResNet-18. On ActivityNet, the performance of the above two settings is close. It indicates that end-to-end training is an effective way of enhancing efficient mid-resolution models.

\noindent(III) The performance gains of end-to-end learning on ActivityNet are smaller than those on THUMOS14. There are two reasons. 1) The performance gain on ActivityNet only reflects \textbf{the effect of end-to-end learning on the localization sub-task}, as the detectors only perform class-agnostic localization on this dataset. 
To verify this, we evaluate the effect of end-to-end training on class-aware detection on ActivityNet. Compared with head-only learning, end-to-end learning enjoys a gain of 5.70\% mAP (19.38\% to 25.08\%, with TSM ResNet50), which is larger than the gain on the localization sub-task. It means the classification sub-task also benefits from E2E learning.
2) \textbf{ActivityNet and THUMOS14 have different characteristics}.
THUMOS14 poses a great challenge to temporal localization, as the actions are shorter and each video has a large amount of background (71\%) on average. Differently, on ActivityNet, actions are much longer and each video has only 36\% background on average. To verify the effect of different characteristics, we conduct a comparison of E2E and head-only learning on HACS Segments~\cite{zhao2019hacs}, which shares the same classes and has a similar distribution as ActivityNet. We observe that E2E learning results in an improvement of 6.28\% mAP (19.28\% to 25.70\%, with TSM ResNet-50), similar to the observation on class-aware detection on ActivityNet.

\begin{table}[]
    \centering
    \begin{tabular}{l|c|c|c}
    \toprule

    Paradigm&Img. Res.&ResNet-18&ResNet-50\\
    \midrule
        \multicolumn{4}{c}{\textbf{THUMOS14}} \\
        \midrule
        Head-only&224$^2$ &33.79&37.77\\
    \midrule 
        Head-only&96$^2$ &28.90&34.04 \\
        E2E&96$^2$ &\textbf{38.31}&\textbf{45.25}\\
        \midrule
        Gain&-&+9.41&+11.21\\
        \bottomrule
        
        \multicolumn{4}{c}{\textbf{ActivityNet}} \\
        \midrule
    %   \multicolumn{4}{c}{TSM ResNet-50}\\
        Head-only&224$^2$ &\textbf{ 33.43}&\textbf{34.21}\\
        \midrule 
        Head-only&96$^2$ &32.12&32.76 \\
        E2E&96$^2$ &33.42&34.14\\
        \midrule
        Gain&-&+1.30&+1.38\\
    \bottomrule
    \end{tabular}
    \caption{\textbf{Head-only learning~\vs~end-to-end (E2E) learning}. Average mAP is reported.   Head: TadTR. Video encoder: TSM. %The results with 224\x224
    }
    \label{tab:finetuning}
\end{table}

\vspace{1ex}\noindent \textbf{Image Augmentations.}
One particular benefit of end-to-end learning is the feasibility of image augmentations.
Except for the commonly used random cropping and random horizontal flipping augmentations, we also study stronger augmentations, including random rotation and random photometric distortion. The effect of these augmentations is depicted in Tab.~\ref{tab:augmentation}.
On both datasets, they result in large performance gains. On THUMOS14, random cropping brings a 3.32\% improvement. Random flipping further improves the performance by 1.09\%. Using stronger augmentations, the average mAP is boosted by 1.35\%. In total, the improvement is 5.76\%. This is reasonable as THUMOS14 is a relatively smaller dataset.
On ActivityNet, the average mAP improves from 31.98\% to 33.42\% (+1.44\%). We find that stronger data augmentations do not provide a clear performance gain, as ActivityNet is already a large-scale dataset. It is worth noting that end-to-end learning without image augmentation performs worse than head-only learning, possibly due to overfitting.

\begin{table}[]
    \centering
    \begin{tabular}{l|cccc}
    \toprule
    Augmentation&\multicolumn{4}{c}{Average mAP}\\
    \midrule
        Cropping        &   &\checkmark&\checkmark&\checkmark\\
        Horizontal Flipping        &   &           &\checkmark&\checkmark\\
        Rotation      &   &           &           &\checkmark\\
        Distortion  &   &           &           &\checkmark\\
        \midrule
        THUMOS14&39.49&42.81&43.90&\textbf{45.25}\\
        ActivityNet&31.98&33.24&33.40&\textbf{33.42} \\
    \bottomrule
    \end{tabular}
        \caption{The effect of \textbf{image augmentations}. Head: TadTR. Video encoder: TSM ResNet-50 on THUMOS14 and TSM ResNet-18 on ActivityNet.}
    \label{tab:augmentation}
\end{table}

\subsection{Evaluation of Design Choices}

\begin{table*}[]
    \centering
    \begin{tabular}{l|ccc|cccc}
    \toprule
    Head&FLOPs/G&Latency/ms&Params&0.5&0.75&0.95&Avg.\\
    \midrule
        AFSD*&249.4/3.3& 145.5/26.9&\textbf{30M}& -&-&-&32.90\\
        G-TAD&  169.2/44.6&  99.5/31.0  &38M& 49.22&34.55&   4.74&33.17     \\
        TadTR&\textbf{125.6}/\textbf{0.9}&\textbf{78.4}/\textbf{9.7} &45M&\textbf{49.56} &\textbf{35.24}& \textbf{9.93}&\textbf{34.35}\\
        % G-TAD&
    \bottomrule
    \end{tabular}
    \caption{Comparison of end-to-end trained detectors with different \textbf{heads} on ActivityNet. Encoder: \textbf{I3D}. All methods use 384 frames inputs (except * uses 768 frames). The values before and after each slash are measured for the full network and the head respectively.}
    % \small
    \label{tab:detection_head_anet}
\end{table*}

\begin{table}[]
    \centering
        \begin{tabular}{l|cccccc}
    \toprule
    Head&0.3&0.4&0.5&0.6&0.7&Avg.\\
    \midrule
    \multicolumn{7}{c}{I3D with a frame rate of 10 FPS}\\ \midrule
    AFSD*&57.7&52.8&45.4&34.9&22.0& 43.6\\
        G-TAD& 52.5&45.9&37.6&28.5&19.1&36.7\\
        TadTR&\textbf{59.6}&\textbf{54.5} &\textbf{47.0}& \textbf{37.8}& \textbf{26.5}& \textbf{45.1}\\
        \midrule
            \multicolumn{7}{c}{TSM ResNet-50 with a frame rate of 2.5 FPS}\\ \midrule
            AFSD&56.0&50.0&42.2&32.8&20.5&40.3\\
            G-TAD&51.5&43.4&33.8&23.5&13.6&33.2 \\
                TadTR&\textbf{58.1}&\textbf{52.9}& \textbf{44.6}& \textbf{36.2}& \textbf{24.1}& \textbf{43.2}\\
    \bottomrule
    \end{tabular}
\caption{Comparison of end-to-end trained detectors with different \textbf{heads} on THUMOS14.
    * Results from ~\cite{lin2021learning}.
    }
    \label{tab:detection_head}
\end{table}

\noindent \textbf{Detection Heads.}
Tab.~\ref{tab:detection_head_anet} and Tab.~\ref{tab:detection_head} 
    compare different heads on ActivityNet and THUMOS14 respectively. 
Note that we use the labels from the external video-level action classifiers for G-TAD following the original paper~\cite{xu2020g}, as this head is designed to generate class-agnostic proposals.
Although the detection head only contributes to a small fraction of the computation cost of a detector, there are still differences between detectors in performance, computation cost, and model size.
To be specific:

\noindent (I) Performance: On both datasets, the query-based detector TadTR achieves the best performance. Its advantage is large in mAP at high IoU thresholds. Specifically, it outperforms G-TAD by 5.19\% at the strict IoU threshold 0.95 on ActivityNet. On THUMOS14, it outperforms AFSD~\cite{lin2021learning} by 4.5\% in terms of mAP@0.7 on THUMOS14 using the I3D encoder.  We observe that G-TAD achieves much lower performance on THUMOS14, as the external action classifier restricts the classification accuracy. Making class-aware predictions like the other two heads is likely to boost its performance.

\noindent (II) Computation cost: G-TAD has much higher FLOPs than the other two heads, as it generates dense anchors. It acounts for around 1/3 of the full network's latency. 
TadTR has the lowest latency as it outputs very sparse detections.
Therefore, reducing the number of detections is a promising direction for building efficient detectors.

\noindent (III) Model size: AFSD has the smallest model size, only 66.7\% that of TadTR. Therefore it is a better choice when a small model size is desired.

\begin{table*}[]
    \centering
    \begin{tabular}{l|ccc|cccccc|cccc}
    \toprule
        \multirow{2}*{Encoder} &\multirow{2}*{FLOPs}&\multirow{2}*{Latency} &\multirow{2}*{Params}&\multicolumn{6}{c|}{THUMOS14}&\multicolumn{4}{c}{ActivityNet}\\
        &&&&0.3&0.4&0.5&0.6&0.7&Avg.&0.5&0.75&0.95&Avg.\\
    \midrule
        TSM R18&\textbf{32.3G}&\textbf{25.7ms}&\textbf{24M}&52.8 &47.9& 39.8& 30.7& 20.3& 38.3 &49.12&34.00&9.74& 33.42\\
        TSM R50&73.2G&41.4ms&36M & 60.5& 55.5& 47.5& 37.6& 25.3& 45.3&49.59& 34.74& 9.72& 34.14\\
        TSN R50&73.2G&41.4ms&36M&44.2& 39.6& 31.9& 22.9& 13.7& 30.5&48.97& 33.26& 7.84 &32.65 \\

        I3D &125.6G&78.4ms&45M&59.6&54.5 &47.0& 37.8& 26.5& 45.1&49.56 &35.24& 9.93&34.35\\

        SF R50&62.1G&63.5ms&46M&\textbf{69.4}&\textbf{64.3}&\textbf{56.0}&\textbf{46.4}&\textbf{34.9}&\textbf{54.2}&\textbf{50.13}& \textbf{35.78}& \textbf{10.52} &\textbf{35.10}\\

    \bottomrule
    \end{tabular}
        \caption{Comparison of end-to-end trained detectors with different \textbf{video encoders}. FLOPs and latency are measured on ActivityNet.}
    \label{tab:encoder_comparison}
\end{table*}

\vspace{1ex}\noindent\textbf{Video  Encoders.}
Tab.~\ref{tab:encoder_comparison} compares different encoders on THUMOS14 and ActivityNet. We observe that:

\noindent(I) While using a smaller backbone reduces the computation cost, it may severely downgrade the detection performance. For example, TSM ResNet-18 achieves 7\% lower average mAP than TSM ResNet-50 on THUMOS14. 

\noindent(II) \textbf{Motion information is important for temporal action detection}. The commonly used TSN encoder falls far behind the others, for lack of motion information modeling. It is even weaker than TSM ResNet-18, which models motion information but has a smaller backbone.

\noindent(III) TSM performs on par with I3D, another typical video encoder in TAD. Meanwhile, its latency is around half of I3D. 
We observe that the advantage of I3D lies in mAP at high IoU thresholds, as it uses a higher sampling frame rate. Therefore TSM is a desirable replacement for I3D when there is no strict demand on localization accuracy.

\noindent(IV) SlowFast achieves the best performance on both datasets. This is reasonable, as SlowFast is a state-of-the-art action recognition model. Its advantage is particularly large on THUMOS14, as the fast pathway can effectively model fast-changing motion, which helps to localize short actions on this dataset. Meanwhile, it is also efficient. It has lower FLOPs than TSM R50, TSN R50, and I3D. The incosistency between FLOPs and latency might be due to the low GPU utilization at low the video resolution.

\begin{figure}
    \centering
    \includegraphics[width=0.85\linewidth]{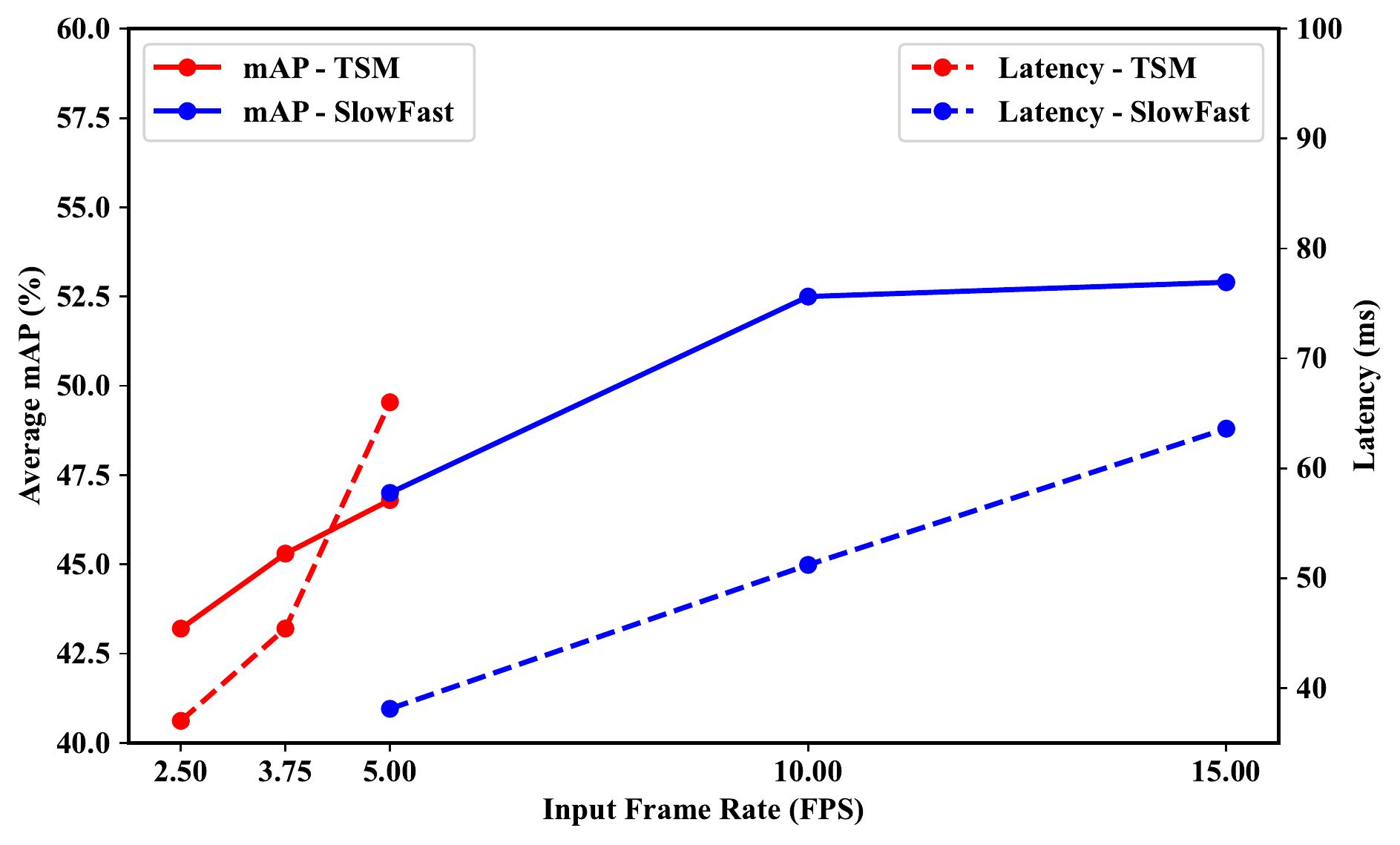}
    \caption{The effect of \textbf{input frame rate} on TAD performance (left Y-axis, solid lines) and latency (right Y-aixs, dashed lines) on THUMOS14. \textcolor{red}{Red} lines and \textcolor{blue}{blue} lines are with TSM ResNet-50 encoder and SlowFast ResNet-50 encoder respectively.}
    \label{fig:resolution_impact}
\end{figure}
\begin{figure}
    \centering
    \includegraphics[width=0.9\linewidth]{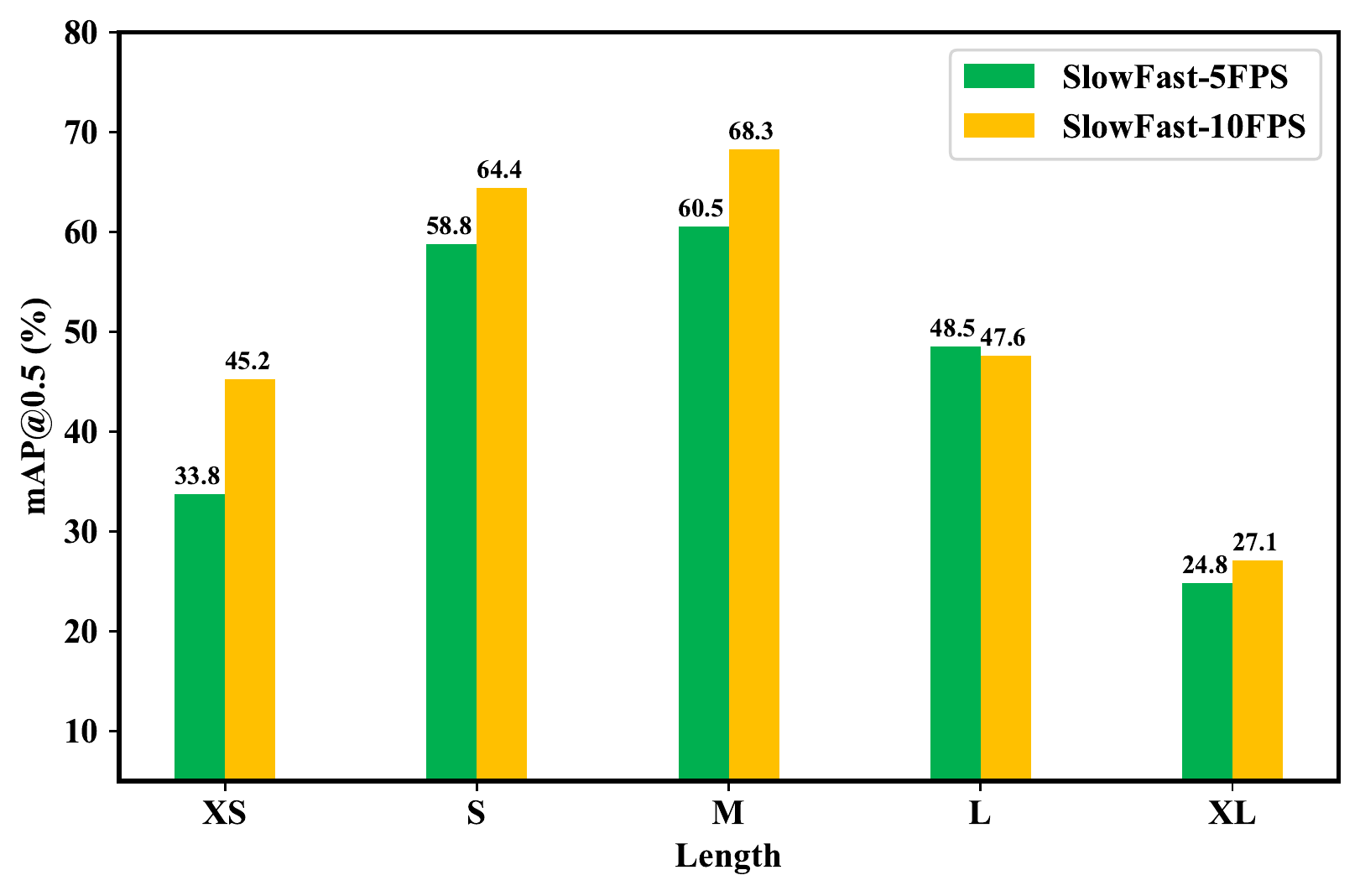}
    \caption{Increasing the input frame rate (from 5 FPS to 10 FPS) helps to detect short actions. Actions are divided into five groups according to their length (in seconds): XS (0, 3], S (3, 6], M (6, 12], L (12, 18], XL (18, inf). Detector: TadTR.}
    \label{fig:backbone_ap_by_length}
\end{figure}

\vspace{1ex} \noindent\textbf{Temporal Resolution.} 
Fig.~\ref{fig:resolution_impact} compares the performance of TadTR using different input frame rates. We use temporal linear interpolation to ensure the output feature sequence has the same length. It is observed that increasing the input frame rate from 2.5 to 5 steadily improves the detection performance of TSM~\cite{lin2019tsm} on THUMOS14, where most actions instances are very short. 
% This reveals that the default input frame rate is insufficient for temporal action detection for short actions. 
Therefore, we switch the encoder to SlowFast~\cite{feichtenhofer2019slowfast}, which performs as well as TSM at 5 FPS but runs much faster, owing to the efficiency of its fast pathway. The performance improves by a sizable margin as the frame rate increases to 10 FPS.  
We show in Fig.~\ref{fig:backbone_ap_by_length} that the increase is mainly from short actions. It indicates that a high frame rate is important for detecting short actions. 
Further increasing the frame rate does not bring a clear performance gain.

\begin{figure}
    \centering
    \includegraphics[width=0.95\linewidth]{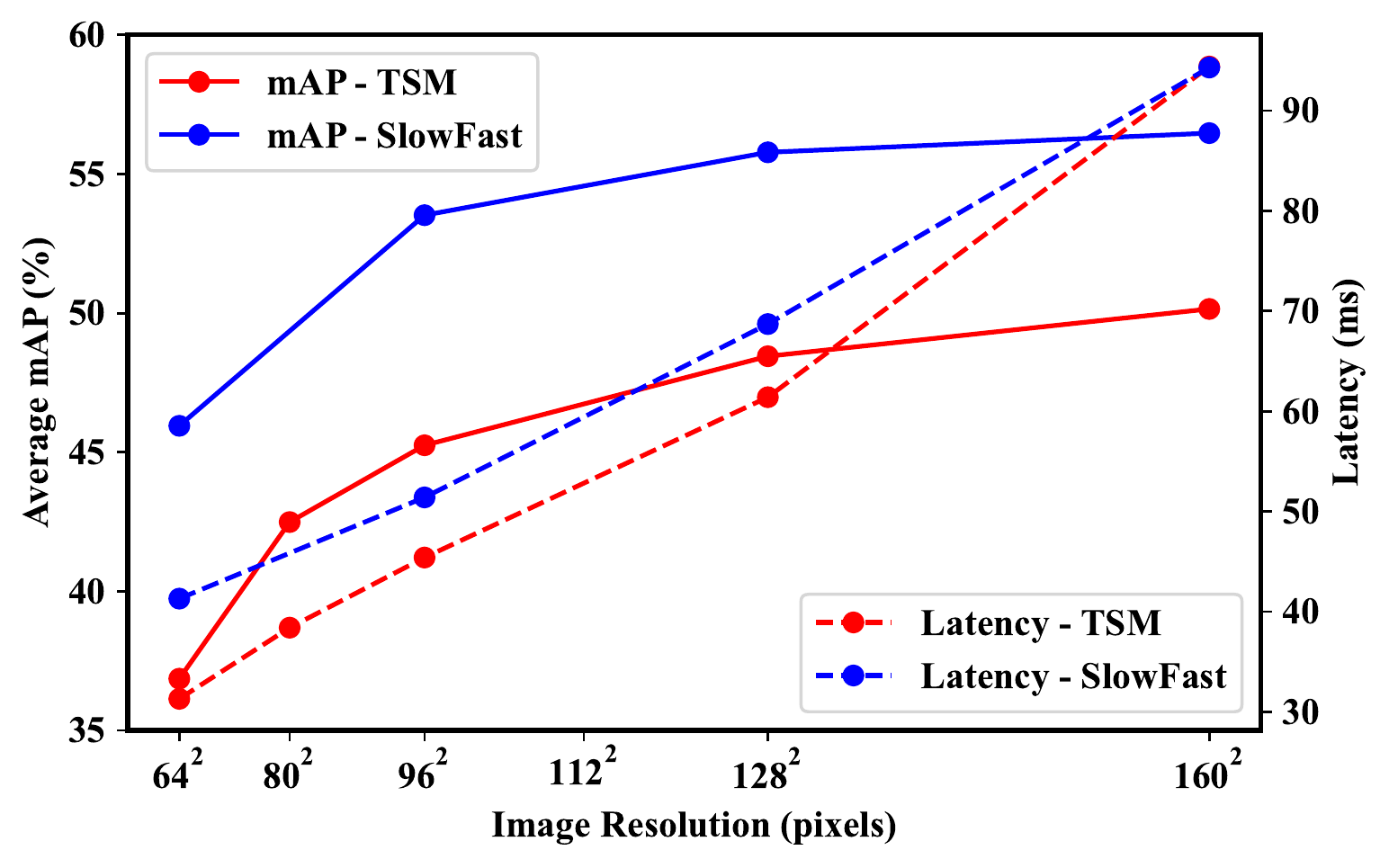}
    \caption{The effect of \textbf{image resolution} on THUMOS14. The input frame rate is set to 3.75 and 10 for TSM and SlowFast respectively. Detector: TadTR. }
    \label{fig:img_size_impact}
\end{figure}

\vspace{1ex} \noindent  \textbf{Image Resolution.}  Fig.~\ref{fig:img_size_impact} compares the performance with different input image resolution on THUMOS14. The slop of each line segment roughly represents the average performance gain per pixel.  We observe that:

\noindent (I) Increasing image resolution boosts TAD performance, at the expense of efficiency. The improvement is especially large when the resolution increases from small ($64^2$) to medium ($96^2$).  It indicates that a sufficient image resolution is critical for good performance. After that, the average performance gain per pixel gradually decreases. Therefore we choose the $64^2$ resolution for a balance between performance and efficiency.

\noindent (II) Increasing image resolution is less important than switching to a more suitable video encoder. We find that SlowFast ResNet-50 encoder with $96^2$ resolution outperforms TSM ResNet-50 encoder with $160^2$ resolution.

\begin{table*}
    \centering
        \begin{tabular}{l|cc|ccccc|>{\columncolor[gray]{0.9}}c|ccc|>{\columncolor[gray]{0.9}}c}
    \toprule
        \multirow{2}{*}{Method} & \multirow{2}{*}{Encoder} & \multirow{2}{*}{Flow} & \multicolumn{6}{c}{THUMOS14} &       \multicolumn{4}{|c}{ActivityNet} \\ \cline{4-13} 
                        &                          &                     & 0.3  & 0.4 & 0.5 & 0.6 & 0.7 & Avg. & 0.5   & 0.75   & 0.95   & Avg.  \\
        \midrule
            Yeung~\etal~\cite{yeung2016end}&VGG16&& 36.0&26.4& 17.1 &-&-&-&-&-&-&-\\
            TAL-Net~\cite{chao2018rethinking}&I3D&\checkmark&53.2&48.5&42.8&33.8&20.8&39.8&38.23&18.30&1.30&20.22\\
    BSN\cite{lin2018bsn}&TSN &\checkmark&53.5&45&36.9&28.4&20&-&46.45&	29.96&8.02&30.03\\
    BMN~\cite{lin2019bmn}& TSN&\checkmark&56.0&47.4&38.8&29.7&20.5&38.5&50.07&34.7&8.29&33.85\\
            G-TAD~\cite{xu2020g} &TSN&\checkmark&54.5&47.6&40.2&30.8&23.4&39.3&50.36& 34.60&9.02 &34.09\\
            BC-GNN~\cite{bai2020boundary}&TSN&\checkmark&57.1&49.1&	40.4&	31.2&	23.1&	40.2&50.56&34.75&9.37&34.26\\
            A2Net~\cite{yang2020revisiting}&I3D&\checkmark&58.6&54.1&45.5&32.5&17.2&41.6&43.55&28.69&3.70&27.75\\
            
        P-GCN~\cite{zeng2019graph}&I3D&\checkmark&63.6&57.8&49.1&-&-&-&48.26&33.16&3.27&31.11\\
        MUSES~\cite{Liu_2021_CVPR}&I3D&\checkmark&68.9&	64.0&\textbf{56.9}&46.3&	31.0&	53.4&50.02&34.97&6.57&33.99\\
        VSGN~\cite{zhao2021video}&TSN&\checkmark&66.7&	60.4&	52.4&	41.0&	30.4&	50.2&52.38&\textbf{36.01}&8.37&35.07\\
        \midrule
        
        % \multicolumn{13}{l}{\textbf{End-to-end trained models}}\\
        S-CNN~\cite{shou2016temporal}&C3D&&36.3&28.7&19&10.3&5.3&-&-&-&-&-\\
        
        R-C3D~\cite{xu2017r}&C3D&&44.8&35.6&28.9&-&-&-&26.80&-&-&-\\
        SS-TAD\cite{buch2017end}&C3D&&	45.7&	-&	29.2&	-&	9.6& -&-&-&-&-\\
            CDC~\cite{shou2017cdc} & C3D& & 40.1& 29.4& 23.3& 13.1 &7.9& 22.8& 45.3& 26.0 &0.2 &23.8\\
            SSN~\cite{zhao2017cuhk}&TSN&\checkmark&51.9&41.0&29.8&-&-&-&-&-&-&-\\
        GTAN\cite{long2019gaussian}& P3D&&57.8&	47.2	&38.8&	-&	-&-&52.61&34.14&8.91&34.31\\	
        PBRNet~\cite{liu2020progressive}&I3D&\checkmark&58.5&	54.6	&51.3	&41.8&	29.5&	47.1&\textbf{53.96}&34.97&8.98&35.01\\
        AFSD~\cite{lin2021learning}&I3D&\checkmark&67.3& 62.4 &55.5& 43.7 &31.1&52.0&52.38&35.27&6.47&34.39\\
        AFSD-RGB\cite{lin2021learning}&I3D&&57.7&52.8&45.4&34.9&22.0&43.6&-&-&-&32.90\\
        Ours & SF R50&&\textbf{69.4}&\textbf{64.3}&56.0&\textbf{46.4}&\textbf{34.9}&\textbf{54.2}&50.47& 35.99 &\textbf{10.83}& \textbf{35.10} \\
        \bottomrule
\end{tabular}
    \caption{State-of-the-art comparison in terms mAP at different thresholds. Only the methods in the second group are end-to-end trained.
%  The ``Flow'' column indicates whether a method uses optical flow.
    }
    \label{tab:comparison_merged}
\end{table*}

Due to space limit, we put the analyses of the the effect of video resolution on ActivityNet \textbf{in the supplementary}. We also analyze the effects of the other two design choices, feature fusion and the frame sampling manner in it.

\subsection{Comparison with State-of-the-art Methods}
% In above studies, we identify an efficient video encoder and a strong action detector that can achieve strong performance with a relative low cost.
In the above study, we identify that SlowFast well balances between performance and accuracy and that TadTR is a strong and efficient action detection head.
Here we combine them as a baseline detector for comparison with state-of-the-art methods. The default resolution is used.

\noindent\textbf{Detection Performance.} 
Tab.~\ref{tab:comparison_merged} compares the detection performance of different methods on THUMOS14 and ActivityNet. 
We divide them into two groups according to whether end-to-end training is used.
Alghough S-CNN~\cite{shou2016temporal}, CDC~\cite{shou2017cdc}, and SSN~\cite{zhao2017temporal} are multi-stage methods, we still regard them as end-to-end methods as the encoder and the head are jointly optimized in each stage. 
We observe that:

\noindent(I) On both datasets, the baseline detector achieves the best performance among end-to-end methods. This is a result of the better video encoder and the stronger detection head.

\noindent(II) Without optical flow, this detector surpasses those two-stream methods that are based on pre-trained features, such as MUSES~\cite{Liu_2021_CVPR} and VSGN~\cite{zhao2021video}. Similarly, AFSD-RGB also outperforms many two-stream methods. It means that \textbf{optical flow is not necessary for TAD}, as the video encoders learn to capture cues of action boundaries from RGB frames via end-to-end training.

\vspace{1ex}\noindent\textbf{Computation Cost.} 
In Tab.~\ref{tab:runtime}, we already compare the computation cost with of the state-of-the-art methods. The baseline detector has a lower computation cost than previous end-to-end detector, as a result of the more efficient video encoder and detection head.
Compared with the state-of-the-art method~\cite{Liu_2021_CVPR} that is based on pre-trained features, the baseline runs 126$\times$ faster. 
We analyze the reason for the huge difference between their computation costs \textbf{in the supplementary}.

Besides, we compare the inference speed in terms of inference FPS in Tab.~\ref{tab:infer_fps}. Note that this metric has a bias. It is more favorable for methods that use a high input frame rate (\eg, 25 in R-C3D~\cite{xu2017r}). Therefore we also report the speedup ratio, \ie the ratio of inference FPS to the input frame rate. Our detector runs at 5076 FPS and has a speedup ratio of 508, which is much faster than the other end-to-end methods.

\begin{table}[]
    \centering
    \begin{tabular}{l|c|cc}
    \toprule
    Model&GPU&Infer. FPS&Speedup\\
    \midrule
        S-CNN~\cite{shou2016temporal}&-&60&-\\
        % DAP~\cite{escorcia2016daps}&-&134\\
        CDC~\cite{shou2017cdc}&TITAN Xm&500&-\\
        SS-TAD~\cite{buch2017end}&TITAN Xm&701&23\\
        R-C3D~\cite{xu2017r}&TITAN Xm&569&23\\
        R-C3D~\cite{xu2017r}&TITAN Xp&1030&45\\
        AFSD~\cite{lin2021learning}&TITAN Xp&3403*&340*\\
        Ours&TITAN Xp&5076&508\\
        % Ours&RTX 3090 & 6531\\
        \bottomrule
    \end{tabular}
    \caption{Comparison of the inference speed, measure by the number of processed frames per second (FPS) and the speedup ratio. *Only measure the RGB network.}
    \label{tab:infer_fps}

\end{table}

\section{Conclusion}
We conduct an empirical study of end-to-end temporal action detection. We show that end-to-end training gives rise to much better performance than the traditional head-only learning paradigm, where the video encoder is only optimized for action recognition. 
We also study multiple factors that affect the performance and accuracy of end-to-end temporal action detection to seek a efficiency-accuracy trade-off. 
Based on our findings, we build a mid-resolution detector that outperforms previous end-to-end methods while running more than 4$\times$ faster. It is also encouraging that the detector surpasses the previous two-stream models without optical flow.
The results show that end-to-end learning is a promising direction for building strong and efficient TAD models.    
Hopefully, this work can serve as a useful reference guide for end-to-end training and inspire future research.

\noindent\textbf{Limitation.} End-to-end learning may still restrict the use of stronger video encoders, higher video resolution due to the constraint of GPU memory. In the future, we plan to explore the complementarity of end-to-end learned features with pre-trained features to address this limitation. 

\noindent\textbf{Acknowledgement.} This work was supported by National Key R\&D Program of China (No. 2018YFB1004600).

% \newpage
\appendix
\setcounter{table}{0}   % start numbering from 0
\setcounter{figure}{0}
\renewcommand{\thetable}{A\arabic{table}}
\renewcommand{\thefigure}{A\arabic{figure}}

\section{Implementation Details}
\noindent\textbf{Initialization of Video Encoders.} All the video encoders are initialized with the pretrianed models on Kinetics-400~\cite{carreira2017quo}. This is similar to the common practice of ImageNet~\cite{deng2009imagenet} pre-training in image understanding.

\vspace{1ex}\noindent\textbf{Combination with External Video Labels on ActivityNet.} As mentioned in the main body of the paper, most videos on ActivityNet only contain one action class. Therefore, most previous works~\cite{lin2018bsn,lin2019bmn,long2019gaussian,zhao2021video,xu2020g,alwassel_2021_tsp,xu2021low,zhao2020bottom,zhu2021enriching,sridhar2021class}, \textbf{including some end-to-end methods}~\cite{liu2020progressive,lin2021learning,shou2017cdc}, decompose temporal action detection (TAD) into class-agnostic temporal localization and video-level action classification.
Following these works, we use the video-level action classification results of~\cite{zhao2017cuhk}, a winning solution in ActivityNet Challenge 2017. To be concrete, we assign the top two video-level classes predicted by~\cite{zhao2017cuhk} to all class-agnostic detections, forming class-aware detections. The confidence score of the original class-agnostic detection and the classification score by~\cite{zhao2017cuhk} are fused by multiplication.

\vspace{1ex}\noindent\textbf{Implementation Details of AFSD.} When implementing  AFSD~\cite{lin2021learning}, we follow the details in their official code. They use a smaller batch size (1~\vs~4), a smaller learning rate ($10^{-5}$~\vs~$10^{-4}$), and a longer training schedule (16 epochs~\vs~12 epochs).  We tried the setting defined in this paper but observed a performance drop of around 1.5\% average mAP. We also observe that using stronger augmentations does not improve performance on AFSD. Therefore, we stick to the original settings. 

\section{Additional Results}
\noindent\textbf{The Effect of Batch Size.} Tab.~\ref{tab:batch_size_effect} studies the effect of batch size for training. We change the learning rate following the linear scaling rule~\cite{goyal2017accurate} when changing the batch size. We observe that increasing batch size from 4 (the default setting) to 16 results in similar performance. Therefore, we may use a larger batch size to improve GPU utilization and speed up training. For example, we can finish training of TadTR~\cite{liu2021end} with TSM ResNet-18~\cite{lin2019tsm} encoder on ActivityNet using two NVIDIA RTX 3090 GPUs in \textbf{41 minutes}. Even so, we stick to a relatively small batch size so that our experiments can be reproduced more easily.

\begin{table}[]
    \centering
    \caption{The effect of batch size, measured by average mAP on ActivityNet. Encoder: \textbf{TSM ResNet-18}. Detector: \textbf{TadTR}. Only cropping and horizontal flipping augmentations are used. All models are trained using a single GPU (except * uses two GPUs). }
    \label{tab:batch_size_effect}
    \begin{tabular}{c|ccc}
    \toprule
         Batch Size &   4   &   8   &  16 \\
         \midrule
         mAP& 33.40 & 33.43 & 33.25\\
         Training Time&96min&85min&41min*\\
         \bottomrule
    \end{tabular}
\end{table}

\vspace{1ex}\noindent\textbf{Generality of the Effectiveness of End-to-End Training.} 
In the main body of this paper, we validate the effectiveness of end-to-end training on TadTR with TSM~\cite{lin2019tsm} encoders. Here we also conduct the validation on more video encoders (SlowFast~\cite{feichtenhofer2019slowfast} and I3D~\cite{carreira2017quo}) and detection heads (AFSD~\cite{lin2021learning} and G-TAD~\cite{xu2020g}). The results are summarized in Tab.~\ref{tab:e2e_effectiveness_extra}. All results are obtained with the default setting. We see that end-to-end learning consistently improves detection performance. The performance gain ranges between 9.79\% and 10.54\% on THUMOS14~\cite{jiang2014thumos}. 
On ActivityNet, the performance gain is at least 1.77\%. The results show that the effectiveness of end-to-end learning is general. Note that the results of G-TAD with TSM ResNet-18 are lower than those in BSP~\cite{xu2021boundary} and LowFi~\cite{xu2021low}. The reason might be the low resolution of videos. 

It is interesting that the performance difference between TadTR and G-TAD increases when we switch from head-only learning to end-to-end learning. It indicates that traditional head-only learning might not be appropriate for benchmarking different approaches, as head-only learning restricts the performance of an approach.

Besides THUMOS14 and ActivityNet, we also evaluate the effect of end-to-end learning on HACS Segments~\cite{zhao2019hacs}. As can be observed in Tab.~\ref{tab:hacs_e2e_effect}, end-to-end training improves the performance by 4.71\% and 6.42\% in terms of average mAP with TSM ResNet-18 and TSM ResNet-50, respectively. This again demonstrates the benefit of E2E learning and its generality.

\begin{table}[tb]
\subfloat[
\textbf{THUMOS14}
\label{tab:th14_e2e_effect_more}
]{
\centering
\begin{minipage}{\linewidth}{\begin{center}
\begin{tabular}{c|*{3}{p{1.7cm}<{\centering}}}
\toprule
    Encoder& I3D & SF R50& TSM R50 \\
    Head& TadTR&TadTR&AFSD\\
\midrule
    Head-only &32.66&44.55&30.52 \\
    E2E&\textbf{45.06}&\textbf{54.17} &\textbf{40.31}\\
    \midrule
    Gain&+10.54&+9.62&+9.79\\
\bottomrule
\end{tabular}
\end{center}}\end{minipage}
}
\vspace{0.5em}
\subfloat[
\textbf{ActivityNet}
\label{tab:anet_e2e_effect_more}
]{
\centering
\begin{minipage}{\linewidth}{\begin{center}
\begin{tabular}{c|*{3}{p{1.7cm}<{\centering}}}
\toprule
    Encoder& \multicolumn{2}{c}{SF R50}& TSM R18 \\
    Head& TadTR&G-TAD&G-TAD\\
\midrule
    Head-only &32.61&32.53&31.12 \\
    E2E&\textbf{35.10}&\textbf{34.36}&\textbf{32.89}\\
    \midrule
    Gain&+2.49&+1.83&+1.77\\
\bottomrule
\end{tabular}
\end{center}}\end{minipage}}
\vspace{-0.2cm}
\caption{End-to-end learning is effective for different video encoders and detection heads. SF: SlowFast. R18/50: ResNet-18/50. Performance measured by average mAP.}
\label{tab:e2e_effectiveness_extra}
\end{table}

\begin{table}[]
    \centering
    \begin{tabular}{c|cccc}
    \toprule
        Paradigm&0.5&0.75&0.95&Avg. (Gain)\\
                 \midrule
        \multicolumn{5}{c}{TSM ResNet-18} \\
        \midrule
         Head-only& 26.29 &16.14& 4.59 &16.56 \\
         E2E& \textbf{33.02}& \textbf{20.66}& \textbf{6.50}& \textbf{21.27} (+4.71)\\
         \midrule
        \multicolumn{5}{c}{TSM ResNet-50} \\
        \midrule
          Head-only& 30.69 &18.94& 5.26 &19.28 \\
         E2E& \textbf{40.32}& \textbf{24.97}& \textbf{7.71}& \textbf{25.70} (+6.42)\\
         \bottomrule
    \end{tabular}
    \caption{The effect of end-to-end learning on HACS Segments. Encoder: TSM ResNet-18/50. Head: TadTR.}
    \label{tab:hacs_e2e_effect}
\end{table}

\begin{figure}
    \centering
    \includegraphics[width=0.9\linewidth]{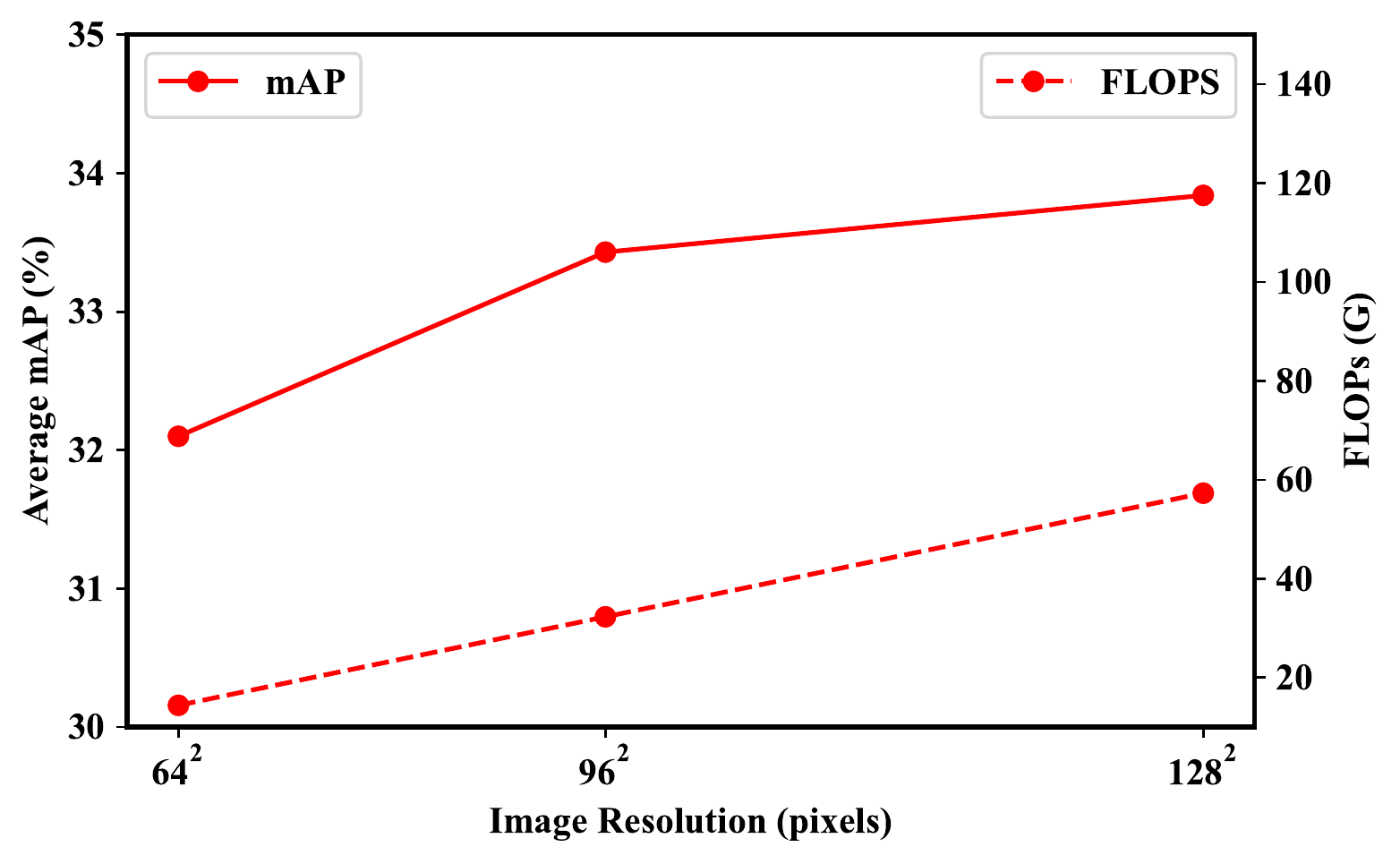}
    \caption{The effect of \textbf{spatial} resolution on ActivityNet. Encoder: \textbf{TSM ResNet-18}. Head: \textbf{TadTR}.}
    \label{fig:anet_spatial_effect}
\end{figure}

\begin{figure}
    \centering
    \includegraphics[width=0.9\linewidth]{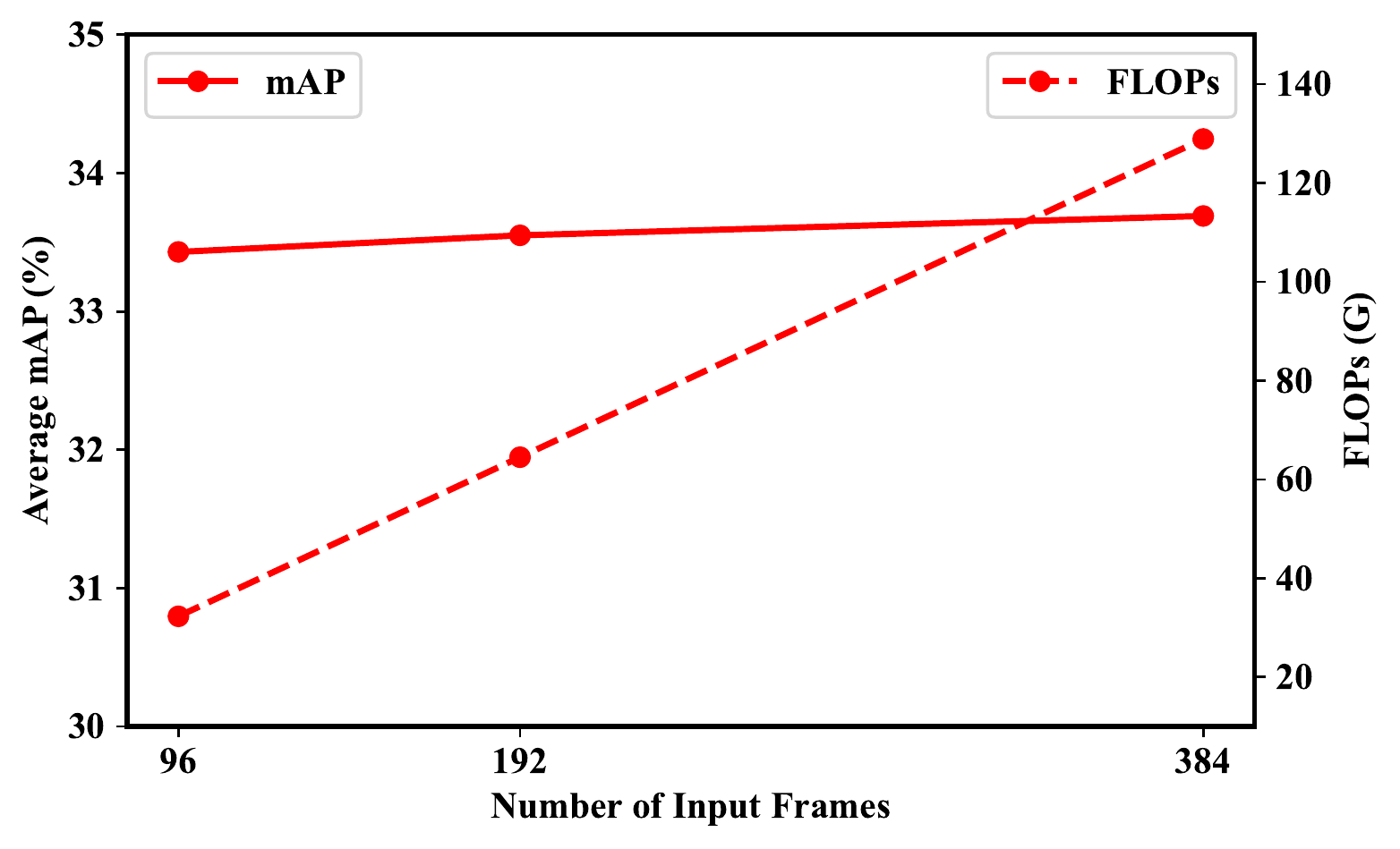}
    \caption{The effect of \textbf{temporal} resolution (number of input frames) on ActivityNet. Encoder: \textbf{TSM ResNet-18}. Head: \textbf{TadTR}.}
    \label{fig:anet_temporal_effect}
\end{figure}

\vspace{1ex}\noindent\textbf{The Effects of Image and Temporal Resolution on ActivityNet.} Fig.~\ref{fig:anet_spatial_effect} and Fig.~\ref{fig:anet_temporal_effect} illustrate the effects of image resolution and temporal resolution (number of input frames) on ActivityNet, respectively. Similar to THUMOS14, increasing image resolution steadily boosts detection performance and also increases computation cost. As the image resolution reaches $96^2$, the performance gain of increasing image resolution decreases. Compared to image resolution, the detection performance is less sensitive to temporal resolution. The reason might be that the action instances on ActivityNet are relatively longer than those on THUMOS14. Besides, the action classes on ActivityNet are more related to scenes than motion. This observation also supports our choice of using sparse frame sampling on ActivityNet.

\vspace{1ex}\noindent\textbf{The Effect of Feature Fusion.} In the I3D encoder, we fuse the features from the fourth stage and fifth stage. We verify the effectiveness of this strategy in Tab.~\ref{tab:feature_fusion_effect}. It is observed that this strategy improves the performance by 3\%. It helps to compensate for the temporal information loss due to a decrease of the frame rate from 20 FPS to 10 FPS. 

\begin{table}

\centering
\begin{tabular}[]{c|c|cc}
    \toprule
   Frame Rate & 20 FPS&\multicolumn{2}{|c}{10 FPS}\\
   \hline
   Feature Fusion & - & - & \checkmark\\
   \midrule
   mAP & 47.5& 42.1&45.1\\ 
%    FLOPs&  \\
   \bottomrule
\end{tabular}
\caption{The effect of multi-scale feature fusion. Encoder: I3D. Detector: TadTR.}
\label{tab:feature_fusion_effect}
\end{table}

\vspace{1ex}\noindent\textbf{The Effect of the Frame Sampling Manner.} By default, the encoder takes all frames from an input clip and extract feature in a temporally fully convolutional manner in our study. An alternative way is to sample fixed-length snippets one by one in a sliding-window manner and extract features for each snippet. The length of snippets is defined when the video encoder is pre-trained for action recognition (\eg, 8 for TSM). It is adopted by most works~\cite{lin2019bmn,xu2020g,liu2021multi} based on offline features. However, it actually increases the total computation cost as adjacent snippets overlap with each other. In end-to-end training, we can still use this manner, at the expense of efficiency. Due to a high memory usage, we are only able to conduct the experiment with TSM ResNet-18 on 4 GPUs. As can be observed in Tab.~\ref{tab:frame_sampling}, the snippet-wise manner actually gives lower performance than the fully convolutional manner, probability due to a limited temporal receptive field.

\begin{table}
    \centering
    \begin{tabular}[tb]{l|cc}
        \toprule
      &    mAP&FLOPs\\
      \midrule
      snippet-wise& 34.25 &171.8G\\
      fully-convolutional  &  36.12 &21.5G  \\
    \bottomrule
    \end{tabular}
    \caption{The effect of frame sampling manner. Encoder: TSM ResNet-18. Detector: TadTR. Dataset: THUMOS14. FLOPs are measured on clips of 25.6 seconds.}
\label{tab:frame_sampling}
\end{table}

\section{Computation Cost Analyses}
\noindent\textbf{Reasons for the Lower Computation Cost Than~\cite{Liu_2021_CVPR}.} In Tab. 1 of the main document, we show that the detector built in this work is 126$\times$ faster (587ms~\vs~74.1s) than the previous state-of-the-art non-end-to-end method~\cite{Liu_2021_CVPR}. The speed-up comes from three aspects. \textbf{Firstly}, we use a smaller image size and a lower frame rate. The setting of image size and frame rate in~\cite{Liu_2021_CVPR} is $224^2$ and $30$ FPS, while the setting in our case is $96^2$ and $10$ FPS. \textbf{Secondly}, we extract features in a fully convolutional manner instead of the conventional snippet-wise manner used in~\cite{Liu_2021_CVPR}. To be concrete, they use a sliding window strategy to sample snippets of 64 frames, which is the default input length of the I3D encoder, for feature extraction. The stride between two adjacent windows is 8 frames, 1/8 of the window length. Therefore, redundant computation is introduced. \textbf{Finally}, the encoder (SlowFast) and the head (TadTR) are more efficient than those in~\cite{Liu_2021_CVPR}. We note that these issues are not unique to~\cite{Liu_2021_CVPR}. They are prevalent in previous methods based on offline features and impede the application of TAD in real-world scenarios. We believe that end-to-end TAD can help eliminate these obstacles.

\begin{table}[]
    \centering
    \begin{tabular}{c|cc|cc}
    \toprule
    Encoder&$N_I$&$N_O$&FLOPs&Latency\\
    \midrule
        TSM R18~\cite{lin2019tsm} & 64&64&\textbf{21.5G}&\textbf{21.2ms}\\
        TSM R50~\cite{lin2019tsm}& 64&64&48.8G&37.0ms \\
        I3D*~\cite{carreira2017quo}&256&64&83.2G&62.1ms\\
        SlowFast R50*~\cite{feichtenhofer2019slowfast}&256&64&41.4G&51.2ms \\
        \midrule
        C3D (VGG-11)~\cite{tran2015learning}&512&64& 906G&277ms\\
        R(2+1)D-18~\cite{tran2018closer}&512&64&960G&382ms\\
        R(2+1)D-34~\cite{tran2018closer}&512&64&1803G&638ms\\
    \bottomrule
    \end{tabular}
        \caption{Comparison of the computation costs of various video encoders, measured on video clips of 25.6 seconds. $N_I$ and $N_O$ are the number of input frames and the length of output features, respectively. Image resolution: $96^2$. *Modified by us to make a temporal output stride of 4.}
    \label{tab:encoder_cost}
\end{table}

\vspace{1ex}\noindent\noindent\textbf{Computation Cost of Various Video Encoders.} Except for the video encoders studied in the main paper, we analyze the computation cost of several video encoders used in previous end-to-end methods~\cite{xu2017r,shou2017cdc} and the pre-training method~\cite{alwassel_2021_tsp}
in Tab.~\ref{tab:encoder_cost}. As they have different settings in temporal pooling, we adjust the number of sampled frames to ensure that they output features of the same length. We observe that C3D~\cite{tran2015learning} and R(2+1)D-18/34~\cite{tran2018closer} are much heavier than SlowFast, although the former two have shallower backbones. For example, C3D, the fastest among them,  is around 5$\times$ slower than SlowFast. They are less appropriate for temporal action detection. Therefore we do not use them in our experiments.

\vspace{1ex}\noindent\textbf{Training Time.} Using the settings described in the implementation details, training SlowFast with TadTR head takes around 4 GPU hours on THUMOS14. On ActivityNet, the training time is around 11 and 1.5 GPU hours using SlowFast and TSM ResNet-18, respectively.

{\small
\bibliographystyle{ieee_fullname}
\bibliography{egbib}

\begin{thebibliography}{10}\itemsep=-1pt

\bibitem{alwassel_2021_tsp}
Humam Alwassel, Silvio Giancola, and Bernard Ghanem.
\newblock Tsp: Temporally-sensitive pretraining of video encoders for
  localization tasks.
\newblock In {\em ICCV Workshops}, 2021.

\bibitem{bai2020boundary}
Yueran Bai, Yingying Wang, Yunhai Tong, Yang Yang, Qiyue Liu, and Junhui Liu.
\newblock Boundary content graph neural network for temporal action proposal
  generation.
\newblock In {\em ECCV}, pages 121--137, 2020.

\bibitem{buch2017end}
Shyamal Buch, Victor Escorcia, Bernard Ghanem, Li Fei-Fei, and Juan~Carlos
  Niebles.
\newblock End-to-end, single-stream temporal action detection in untrimmed
  videos.
\newblock In {\em BMVC}, 2017.

\bibitem{buch2017sst}
Shyamal Buch, Victor Escorcia, Chuanqi Shen, Bernard Ghanem, and Juan~Carlos
  Niebles.
\newblock Sst: Single-stream temporal action proposals.
\newblock In {\em CVPR}, pages 6373--6382, 2017.

\bibitem{caba2015activitynet}
Fabian Caba~Heilbron, Victor Escorcia, Bernard Ghanem, and Juan Carlos~Niebles.
\newblock Activitynet: A large-scale video benchmark for human activity
  understanding.
\newblock In {\em CVPR}, pages 961--970, 2015.

\bibitem{carion2020end}
Nicolas Carion, Francisco Massa, Gabriel Synnaeve, Nicolas Usunier, Alexander
  Kirillov, and Sergey Zagoruyko.
\newblock End-to-end object detection with transformers.
\newblock In {\em ECCV}, pages 213--229, 2020.

\bibitem{carreira2017quo}
Joao Carreira and Andrew Zisserman.
\newblock Quo vadis, action recognition? a new model and the kinetics dataset.
\newblock In {\em CVPR}, pages 4724--4733, 2017.

\bibitem{chao2018rethinking}
Yu-Wei Chao, Sudheendra Vijayanarasimhan, Bryan Seybold, David~A Ross, Jia
  Deng, and Rahul Sukthankar.
\newblock Rethinking the faster r-cnn architecture for temporal action
  localization.
\newblock In {\em CVPR}, pages 1130--1139, 2018.

\bibitem{deng2009imagenet}
Jia Deng, Wei Dong, Richard Socher, Li-Jia Li, Kai Li, and Li Fei-Fei.
\newblock Imagenet: A large-scale hierarchical image database.
\newblock In {\em CVPR}, pages 248--255. Ieee, 2009.

\bibitem{escorcia2016daps}
Victor Escorcia, Fabian~Caba Heilbron, Juan~Carlos Niebles, and Bernard Ghanem.
\newblock Daps: Deep action proposals for action understanding.
\newblock In {\em ECCV}, pages 768--784, 2016.

\bibitem{feichtenhofer2019slowfast}
Christoph Feichtenhofer, Haoqi Fan, Jitendra Malik, and Kaiming He.
\newblock Slowfast networks for video recognition.
\newblock In {\em ICCV}, pages 6202--6211, 2019.

\bibitem{gao2018ctap}
Jiyang Gao, Kan Chen, and Ram Nevatia.
\newblock Ctap: Complementary temporal action proposal generation.
\newblock In {\em ECCV}, September 2018.

\bibitem{gao2017turn}
Jiyang Gao, Zhenheng Yang, Chen Sun, Kan Chen, and Ram Nevatia.
\newblock Turn tap: Temporal unit regression network for temporal action
  proposals.
\newblock In {\em ICCV}, pages 3648--3656, 2017.

\bibitem{Girshick_2015_ICCV}
Ross Girshick.
\newblock Fast r-cnn.
\newblock In {\em ICCV}, December 2015.

\bibitem{goyal2017accurate}
Priya Goyal, Piotr Doll{\'a}r, Ross Girshick, Pieter Noordhuis, Lukasz
  Wesolowski, Aapo Kyrola, Andrew Tulloch, Yangqing Jia, and Kaiming He.
\newblock Accurate, large minibatch sgd: Training imagenet in 1 hour.
\newblock {\em arXiv preprint arXiv:1706.02677}, 2017.

\bibitem{he2017mask}
Kaiming He, Georgia Gkioxari, Piotr Doll{\'a}r, and Ross Girshick.
\newblock Mask r-cnn.
\newblock In {\em ICCV}, pages 2961--2969, 2017.

\bibitem{DBLP:conf/icml/IoffeS15}
Sergey Ioffe and Christian Szegedy.
\newblock Batch normalization: Accelerating deep network training by reducing
  internal covariate shift.
\newblock In {\em ICML}, pages 448--456, 2015.

\bibitem{jiang2014thumos}
YG Jiang, Jingen Liu, A~Roshan Zamir, G Toderici, I Laptev, Mubarak Shah, and
  Rahul Sukthankar.
\newblock Thumos challenge: Action recognition with a large number of classes,
  2014.

\bibitem{kingma2015adam}
Diederik~P. Kingma and Jimmy Ba.
\newblock Adam: {A} method for stochastic optimization.
\newblock In {\em ICLR}, 2015.

\bibitem{li2021three}
Zhihui Li and Lina Yao.
\newblock Three birds with one stone: Multi-task temporal action detection via
  recycling temporal annotations.
\newblock In {\em CVPR}, pages 4751--4760, 2021.

\bibitem{lin2021learning}
Chuming Lin, Chengming Xu, Donghao Luo, Yabiao Wang, Ying Tai, Chengjie Wang,
  Jilin Li, Feiyue Huang, and Yanwei Fu.
\newblock Learning salient boundary feature for anchor-free temporal action
  localization.
\newblock In {\em CVPR}, pages 3320--3329, 2021.

\bibitem{lin2019tsm}
Ji Lin, Chuang Gan, and Song Han.
\newblock Tsm: Temporal shift module for efficient video understanding.
\newblock In {\em ICCV}, pages 7083--7093, 2019.

\bibitem{lin2019bmn}
Tianwei Lin, Xiao Liu, Xin Li, Errui Ding, and Shilei Wen.
\newblock Bmn: Boundary-matching network for temporal action proposal
  generation.
\newblock In {\em ICCV}, pages 3889--3898, 2019.

\bibitem{lin2017single}
Tianwei Lin, Xu Zhao, and Zheng Shou.
\newblock Single shot temporal action detection.
\newblock In {\em ACM MM}, pages 988--996, 2017.

\bibitem{lin2018bsn}
Tianwei Lin, Xu Zhao, Haisheng Su, Chongjing Wang, and Ming Yang.
\newblock Bsn: Boundary sensitive network for temporal action proposal
  generation.
\newblock In {\em ECCV}, September 2018.

\bibitem{liu2020progressive}
Qinying Liu and Zilei Wang.
\newblock Progressive boundary refinement network for temporal action
  detection.
\newblock In {\em AAAI}, volume~34, pages 11612--11619, 2020.

\bibitem{liu2016ssd}
Wei Liu, Dragomir Anguelov, Dumitru Erhan, Christian Szegedy, Scott Reed,
  Cheng-Yang Fu, and Alexander~C Berg.
\newblock Ssd: Single shot multibox detector.
\newblock In {\em ECCV}, pages 21--37. Springer, 2016.

\bibitem{liu2021multi}
Xiaolong Liu, Yao Hu, Song Bai, Fei Ding, Xiang Bai, and Philip~HS Torr.
\newblock Multi-shot temporal event localization: a benchmark.
\newblock In {\em CVPR}, pages 12596--12606, 2021.

\bibitem{Liu_2021_CVPR}
Xiaolong Liu, Yao Hu, Song Bai, Fei Ding, Xiang Bai, and Philip H.~S. Torr.
\newblock Multi-shot temporal event localization: A benchmark.
\newblock In {\em CVPR}, pages 12596--12606, June 2021.

\bibitem{liu2021end}
Xiaolong Liu, Qimeng Wang, Yao Hu, Xu Tang, Song Bai, and Xiang Bai.
\newblock End-to-end temporal action detection with transformer.
\newblock {\em arXiv preprint arXiv:2106.10271}, 2021.

\bibitem{liu2019multi}
Yuan Liu, Lin Ma, Yifeng Zhang, Wei Liu, and Shih-Fu Chang.
\newblock Multi-granularity generator for temporal action proposal.
\newblock In {\em CVPR}, pages 3604--3613, 2019.

\bibitem{long2019gaussian}
Fuchen Long, Ting Yao, Zhaofan Qiu, Xinmei Tian, Jiebo Luo, and Tao Mei.
\newblock Gaussian temporal awareness networks for action localization.
\newblock In {\em CVPR}, pages 344--353, 2019.

\bibitem{piergiovanni2019temporal}
AJ Piergiovanni and Michael~S. Ryoo.
\newblock Temporal gaussian mixture layer for videos.
\newblock In {\em ICML}, 2019.

\bibitem{qing2021temporal}
Zhiwu Qing, Haisheng Su, Weihao Gan, Dongliang Wang, Wei Wu, Xiang Wang, Yu
  Qiao, Junjie Yan, Changxin Gao, and Nong Sang.
\newblock Temporal context aggregation network for temporal action proposal
  refinement.
\newblock In {\em CVPR}, pages 485--494, 2021.

\bibitem{qiu2020borderdet}
Han Qiu, Yuchen Ma, Zeming Li, Songtao Liu, and Jian Sun.
\newblock Borderdet: Border feature for dense object detection.
\newblock In {\em ECCV}, pages 549--564. Springer, 2020.

\bibitem{qiu2017learning}
Zhaofan Qiu, Ting Yao, and Tao Mei.
\newblock Learning spatio-temporal representation with pseudo-3d residual
  networks.
\newblock In {\em ICCV}, pages 5533--5541, 2017.

\bibitem{ren2015faster}
Shaoqing Ren, Kaiming He, Ross Girshick, and Jian Sun.
\newblock Faster r-cnn: Towards real-time object detection with region proposal
  networks.
\newblock In {\em NIPS}, pages 91--99, 2015.

\bibitem{shou2017cdc}
Zheng Shou, Jonathan Chan, Alireza Zareian, Kazuyuki Miyazawa, and Shih-Fu
  Chang.
\newblock Cdc: Convolutional-de-convolutional networks for precise temporal
  action localization in untrimmed videos.
\newblock In {\em ICCV}, pages 1417--1426, 2017.

\bibitem{shou2016temporal}
Zheng Shou, Dongang Wang, and Shih-Fu Chang.
\newblock Temporal action localization in untrimmed videos via multi-stage
  cnns.
\newblock In {\em CVPR}, pages 1049--1058, 2016.

\bibitem{simonyan2014two}
Karen Simonyan and Andrew Zisserman.
\newblock Two-stream convolutional networks for action recognition in videos.
\newblock In {\em NIPS}, pages 568--576, 2014.

\bibitem{sridhar2021class}
Deepak Sridhar, Niamul Quader, Srikanth Muralidharan, Yaoxin Li, Peng Dai, and
  Juwei Lu.
\newblock Class semantics-based attention for action detection.
\newblock In {\em ICCV}, pages 13739--13748, 2021.

\bibitem{tan2021relaxed}
Jing Tan, Jiaqi Tang, Limin Wang, and Gangshan Wu.
\newblock Relaxed transformer decoders for direct action proposal generation.
\newblock In {\em ICCV}, pages 13526--13535, October 2021.

\bibitem{tian2019fcos}
Zhi Tian, Chunhua Shen, Hao Chen, and Tong He.
\newblock Fcos: Fully convolutional one-stage object detection.
\newblock In {\em CVPR}, pages 9627--9636, 2019.

\bibitem{tran2015learning}
Du Tran, Lubomir Bourdev, Rob Fergus, Lorenzo Torresani, and Manohar Paluri.
\newblock Learning spatiotemporal features with 3d convolutional networks.
\newblock In {\em ICCV}, pages 4489--4497, 2015.

\bibitem{tran2018closer}
Du Tran, Heng Wang, Lorenzo Torresani, Jamie Ray, Yann LeCun, and Manohar
  Paluri.
\newblock A closer look at spatiotemporal convolutions for action recognition.
\newblock In {\em CVPR}, pages 6450--6459, 2018.

\bibitem{vaswani2017attention}
Ashish Vaswani, Noam Shazeer, Niki Parmar, Jakob Uszkoreit, Llion Jones,
  Aidan~N. Gomez, Lukasz Kaiser, and Illia Polosukhin.
\newblock Attention is all you need.
\newblock In {\em NIPS}, pages 5998--6008, 2017.

\bibitem{wang2016temporal}
Limin Wang, Yuanjun Xiong, Zhe Wang, Yu Qiao, Dahua Lin, Xiaoou Tang, and Luc
  Van~Gool.
\newblock Temporal segment networks: Towards good practices for deep action
  recognition.
\newblock In {\em ECCV}, pages 20--36, 2016.

\bibitem{xie2018rethinking}
Saining Xie, Chen Sun, Jonathan Huang, Zhuowen Tu, and Kevin Murphy.
\newblock Rethinking spatiotemporal feature learning: Speed-accuracy trade-offs
  in video classification.
\newblock In {\em ECCV}, pages 305--321, 2018.

\bibitem{xu2017r}
Huijuan Xu, Abir Das, and Kate Saenko.
\newblock R-c3d: region convolutional 3d network for temporal activity
  detection.
\newblock In {\em ICCV}, pages 5794--5803, 2017.

\bibitem{xu2021boundary}
Mengmeng Xu, Juan-Manuel P{\'e}rez-R{\'u}a, Victor Escorcia, Brais Martinez,
  Xiatian Zhu, Li Zhang, Bernard Ghanem, and Tao Xiang.
\newblock Boundary-sensitive pre-training for temporal localization in videos.
\newblock In {\em ICCV}, pages 7220--7230, 2021.

\bibitem{xu2021low}
Mengmeng Xu, Juan-Manuel Perez-Rua, Xiatian Zhu, Bernard Ghanem, and Brais
  Martinez.
\newblock Low-fidelity video encoder optimization for temporal action
  localization.
\newblock In {\em NeurIPS}, 2021.

\bibitem{xu2020g}
Mengmeng Xu, Chen Zhao, David~S Rojas, Ali Thabet, and Bernard Ghanem.
\newblock {G-TAD}: Sub-graph localization for temporal action detection.
\newblock In {\em CVPR}, pages 10156--10165, 2020.

\bibitem{yang2021background}
Le Yang, Junwei Han, Tao Zhao, Tianwei Lin, Dingwen Zhang, and Jianxin Chen.
\newblock Background-click supervision for temporal action localization.
\newblock {\em TPAMI}, 2021.

\bibitem{yang2020revisiting}
Le Yang, Houwen Peng, Dingwen Zhang, Jianlong Fu, and Junwei Han.
\newblock Revisiting anchor mechanisms for temporal action localization.
\newblock {\em IEEE Transactions on Image Processing}, 29:8535--8548, 2020.

\bibitem{yeung2016end}
Serena Yeung, Olga Russakovsky, Greg Mori, and Li Fei-Fei.
\newblock End-to-end learning of action detection from frame glimpses in
  videos.
\newblock In {\em CVPR}, pages 2678--2687, 2016.

\bibitem{yuan2017temporal}
Ze-Huan Yuan, Jonathan~C Stroud, Tong Lu, and Jia Deng.
\newblock Temporal action localization by structured maximal sums.
\newblock In {\em CVPR}, volume~2, page~7, 2017.

\bibitem{zeng2019graph}
Runhao Zeng, Wenbing Huang, Mingkui Tan, Yu Rong, Peilin Zhao, Junzhou Huang,
  and Chuang Gan.
\newblock Graph convolutional networks for temporal action localization.
\newblock In {\em ICCV}, pages 7094--7103, 2019.

\bibitem{zhao2021video}
Chen Zhao, Ali~K Thabet, and Bernard Ghanem.
\newblock Video self-stitching graph network for temporal action localization.
\newblock In {\em ICCV}, pages 13658--13667, 2021.

\bibitem{zhao2019hacs}
Hang Zhao, Antonio Torralba, Lorenzo Torresani, and Zhicheng Yan.
\newblock {HACS:} human action clips and segments dataset for recognition and
  temporal localization.
\newblock In {\em ICCV}, pages 8667--8677, 2019.

\bibitem{zhao2020bottom}
Peisen Zhao, Lingxi Xie, Chen Ju, Ya Zhang, Yanfeng Wang, and Qi Tian.
\newblock Bottom-up temporal action localization with mutual regularization.
\newblock In {\em ECCV}, 2020.

\bibitem{zhao2017temporal}
Yue Zhao, Yuanjun Xiong, Limin Wang, Zhirong Wu, Xiaoou Tang, and Dahua Lin.
\newblock Temporal action detection with structured segment networks.
\newblock {\em ICCV}, pages 2914--2923, 2017.

\bibitem{zhao2017cuhk}
Yue Zhao, Bowen Zhang, Zhirong Wu, Shuo Yang, Lei Zhou, Sijie Yan, Limin Wang,
  Yuanjun Xiong, Wang Yali, Dahua Lin, Yu Qiao, and Xiaoou Tang.
\newblock {CUHK} \& {ETHZ} \& {SIAT} submission to {ActivityNet} challenge
  2017.
\newblock {\em arXiv preprint arXiv:1710.08011}, pages 20--24, 2017.

\bibitem{zhu2021enriching}
Zixin Zhu, Wei Tang, Le Wang, Nanning Zheng, and Gang Hua.
\newblock Enriching local and global contexts for temporal action localization.
\newblock In {\em ICCV}, pages 13516--13525, 2021.

\end{thebibliography}
}

\end{document}